\documentclass{article}

\usepackage{arxiv}

\usepackage[utf8]{inputenc} 
\usepackage[T1]{fontenc}    
\usepackage{hyperref}       
\usepackage{url}            
\usepackage{booktabs}       
\usepackage{amsfonts}       
\usepackage{nicefrac}       
\usepackage{microtype}      

\title{Simple Natural Language Processing Tools \\ for Danish}

\author{
  Leon Derczynski\\
  Department of Computer Science\\
  ITU Copenhagen\\
  Denmark DK2300\\
  \texttt{ld@itu.dk} \\
}

\begin{document}
\maketitle

\begin{abstract}
This technical note describes a set of baseline tools for automatic processing of Danish text. 
The tools are machine-learning based, using natural language processing models trained over previously annotated documents.
They are maintained at ITU Copenhagen and will always be freely available.
\end{abstract}


\section{Introduction}
Danish, a language with few resources for automatic processing, is spoken by six million people, largely concentrated in the Scandinavian country of Denmark.
At the NLP (natural language processing) group at ITU Copenhagen, one of our foci is Nordic and Danish languages.
Thus, we aim to include local languages in our NLP research whenever possible, and to provide tools for others working on these languages.
Despite a modest number of researcher in the country, there has been a gap in usable Danish NLP tools that are available to the wider community.
As in every time that new technology is not effectively mediated to those who can use it, this restriction stymies the impact and development NLP in Denmark.
To bridge this artificial lacuna, a basic set of easy-to-run utilities that build on existing datasets and open tools has been developed at ITU, which this note introduces.

The tools are available at \url{https://nlp.itu.dk/resources/}.

\section{daner: Named Entity Recognition}

Named entity recognition means finding words in text that refer to things by a specific name.
This could be e.g. using ``Nanna" to refer to a specific individual, instead of ``them" or ``she" or ``the person"; or referring to ``Moscow" instead of ``the nearest city".
The important thing is that one mentions a name specifically.
Named entity recognition (NER) finds those names. In the case of {\tt daner}, names of Locations, Organisation and Persons are tagged.

The {\tt daner} tool wraps the CoreNLP~\cite{manning2014stanford} named entity recognition component~\cite{finkel2005incorporating}, using the DKIE data~\cite{derczynski2014dkie} developed as part of the GATE tool~\cite{cunningham2014developing}, which is derived from the UD part of the Copenhagen Dependency Treebank~\cite{johannsen2015universal}, itself including data from the Danish Dependency Treebank~\cite{kromann2003danish}.
Further ad-hoc data is added as required, from newswire and other sources.
The data overall comprises 20K tokens from the CDT (largely 1990s newswire), 3K tokens from non-newswire Danish, and 10K tokens from post-2015 newswire.
The mode performs best over Danish newswire text.
The data is annotated for three classes, person (PER), location (LOC), and organisation (ORG).
Tags are in BIO format.

The tool recognises names of people, of organisations, and of locations.
It performs automatic tokenization, and outputs in slashed-tag format.
For example:

\begin{quote}
    En/O stor/O reform/O skal/O derfor/O blandt/O andet/O styrke/O tilliden/O til/O politikere/O og/O medier/O ,/O genopbygge/O tilliden/O til/O Skat/O og/O mindske/O de/O økonomiske/O forskelle/O i/O Danmark/B-LOC ./O 
\end{quote}

Here, ``Danmark" is tagged as a location. The {\tt B} indicates that ``Danmark" is the first token in the location name (the {\underline beginning }).

Additional data is added to an internal repository that is also used to train {\tt daner}; some of this data has copyright restrictions, and so cannot be distributed, though the aggregated statistical representation in {\tt daner}'s model is free from that restriction, and so can be downloaded by anyone.

The NER model is configured to use contextual features, including n-grams up to 5-grams, and to use type sequences and word shape features.

{\tt daner} uses Brown clusters induced over large Danish-language corpus. These are created using the Generalised Brown formulation~\cite{derczynski2016generalised}, which offers more flexibility in use than the classical version -- specifically, one clustering run can be used to generate clusterings of any size. For {\tt daner}, 2500 clusters\footnote{A size that works well for NER in English~\cite{derczynski2015tune}} are induced from a clustering with window size of 5000, over a set of 134M Danish tokens. 
These 134M tokens are taken from Danish Wikipedia and Danish content from the Common Crawl corpus filtered through the FastText language ID~\cite{joulin2016bag}, both using June 2019 data.
The clusters are used inside the model, but also available separately at \url{http://itu.dk/~leod/dansk-brown.tar.bz2}.


\section{dapipe: Tokenization, Part-of-speech tagging, and Dependency Parsing}

Another utility in the toolkit, {\tt dapipe}, provides three functions: breaking text files into words and sentences, finding what class each word is through PoS tagging, and determining one part of the syntax that links a sentence's words together, through dependency parsing.

Defining where words and sentences start and stop is very important if we are going to do NLP.
These processes are both called tokenization.
{\tt dapipe} uses the Universal dependencies method for tokenizing, breaking text into sentences and words.

The next step is to give a ``part of speech" to each word.
This would be something like a noun (substantiv), verb (verbum), or preposition (præposition).
This tells us the function each particular word has.
{\tt dapipe} works generally well, though for social media text, I recommend {\tt structbilty},\footnote{See \url{https://github.com/bplank/bilstm-aux}} which is more resilient~\cite{plank-agic:2018,plank:ea:2016} to the unusual words and noise found in this colloquial setting~\cite{derczynski2015analysis}.

Once the words and their classes are known, one may continue by parsing the words in the sentence.
The {\tt dapipe} tool uses the Universal Dependencies schema\footnote{See \url{https://universaldependencies.org/introduction.html}} for this, finding the main grammatical relations between each word to construct a dependency tree that links every word in the sentence together.

All these tasks are achieved by using UDPipe~\cite{straka2016udpipe} based on the universal dependencies~\cite{nivre2016universal} data for Danish.
Sample output follows.

\small
\begin{verbatim}
leon@blade:~/dapipe$ ./dapipe dktest.txt 

...

# text = Derfor presser EU-siden hårdt på, for at amerikanerne ikke saboterer aftalen, der
blandt andet lader udenlandske observatører inspicere iranske atomanlæg.
1	Derfor	derfor	ADV	_	_	2	advmod	_	_
2	presser	presse	VERB	_	Mood=Ind|Tense=Pres|VerbForm=Fin|Voice=Act	0	root	_	_
3	EU-siden	EU-sid	NOUN	_	Definite=Def|Gender=Com|Number=Sing	2	nsubj	_	_
4	hårdt	hårdt	ADV	_	Degree=Pos	2	advmod	_	_
5	på	på	ADP	_	AdpType=Prep	4	case	_	SpaceAfter=No
6	,	,	PUNCT	_	_	2	punct	_	_
7	for	for	ADP	_	AdpType=Prep	11	mark	_	_
8	at	at	SCONJ	_	_	11	mark	_	_
9	amerikanerne	amerikaner	NOUN	_	Definite=Def|Gender=Com|Number=Plur	11	nsubj	_	_
10	ikke	ikke	ADV	_	_	11	advmod	_	_
11	saboterer	sabotere	VERB	_	Mood=Ind|Tense=Pres|VerbForm=Fin|Voice=Act	2	advcl	_	_
12	aftalen	aftale	NOUN	_	Definite=Def|Gender=Com|Number=Sing	11	obj	_	SpaceAfter=No
13	,	,	PUNCT	_	_	12	punct	_	_
14	der	der	PRON	_	PartType=Inf	17	nsubj	_	_
15	blandt	blandt	ADP	_	AdpType=Prep	17	advmod	_	_
16	andet	anden	PRON	_	Gender=Neut|Number=Sing|PronType=Ind	15	fixed	_	_
17	lader	lade	VERB	_	Mood=Ind|Tense=Pres|VerbForm=Fin|Voice=Act	12	acl:relcl	_	_
18	udenlandske	udenlandsk	ADJ	_	Degree=Pos|Number=Plur	19	amod	_	_
19	observatører	observatør	NOUN	_	Definite=Ind|Gender=Com|Number=Plur	17	obj	_	_
20	inspicere	inspicere	ADV	_	Degree=Cmp	21	advmod	_	_
21	iranske	iransk	ADJ	_	Degree=Pos|Number=Plur	22	amod	_	_
22	atomanlæg	atomanlæg	NOUN	_	Definite=Ind|Gender=Neut|Number=Plur	17	obj	_	SpaceAfter=No
23	.	.	PUNCT	_	_	2	punct	_	SpacesAfter=\n
\end{verbatim}
\normalsize

\section{Summary}

This note describes a free toolkit for basic natural language processing. 
The two tools {\tt daner} and {\tt dapipe} are built on freely-available resources and maintaned at ITU Copenhagen.

Feedback, requests, and so on can be addressed to the ITU NLP research group: \url{http://nlp.itu.dk/}

\bibliographystyle{unsrt}  
\bibliography{references}  

\end{document}